\documentclass{article}

\usepackage{microtype}
\usepackage{graphicx}
\usepackage{subcaption}
\usepackage{booktabs} % for professional tables
\usepackage{url}

\usepackage{hyperref}

\usepackage[preprint]{icml2026}

\usepackage{amsmath}
\usepackage{amssymb}
\usepackage{mathtools}
\usepackage{amsthm}
\usepackage{multirow}

\usepackage{enumitem} % 用于优化列表间距
\usepackage[most]{tcolorbox}

\newtcolorbox{promptbox}[2][]{
    colback=gray!5!white,      % 背景色
    colframe=green!40!black,    % 边框颜色（可以改为你喜欢的颜色）
    fonttitle=\bfseries,        % 标题加粗
    title=#2,                   % 标题内容
    arc=2mm,                    % 圆角幅度
    boxrule=0.5pt,              % 边框粗细
    left=2mm, right=2mm, top=2mm, bottom=2mm, % 内边距
    enhanced,
    attach boxed title to top left={xshift=3mm, yshift=-2mm}, % 标题盒子的位置
    boxed title style={colback=green!40!black}, % 标题盒子背景
    breakable,                           % 允许跨页（适合长案例）
    #1
}

\usepackage[capitalize,noabbrev]{cleveref}

\theoremstyle{plain}

\theoremstyle{definition}

\theoremstyle{remark}

\usepackage[textsize=tiny]{todonotes}

\icmltitlerunning{Submission and Formatting Instructions for ICML 2026}

\begin{document}

\twocolumn[
  \icmltitle{Securing the Floor and Raising the Ceiling: A Merging-based Paradigm for Multi-modal Search Agents}



% 定义共同一作符号（如果需要）
\icmlsetsymbol{equal}{*}

\begin{icmlauthorlist}
  \icmlauthor{Zhixiang Wang}{ant,pku} 
  \icmlauthor{Jingxuan Xu}{ant}
  \icmlauthor{Dajun Chen}{ant}
  \icmlauthor{Yunfang Wu}{pku}
  \icmlauthor{Wei Jiang}{ant}
  \icmlauthor{Yong Li}{ant}
\end{icmlauthorlist}

% 2. 机构定义
\icmlaffiliation{ant}{Ant Group, Hangzhou, China}
\icmlaffiliation{pku}{Peking University, Beijing, China}

% 3. 通讯作者（通常只有一个，或者在名字后加对应标注）
\icmlcorrespondingauthor{Yong Li}{liyong.liy@antgroup.com}

  \icmlkeywords{Machine Learning, ICML}

  \vskip 0.3in
]



\printAffiliationsAndNotice{}  % no special notice (required even if empty)

\begin{abstract}
Recent advances in Vision-Language Models (VLMs) have motivated the development of multi-modal search agents that can actively invoke external search tools and integrate retrieved evidence through multi-step reasoning. While promising, existing approaches typically rely on large-scale supervised trajectories or expensive reinforcement learning (RL), leading to high training cost, instability, and a severe cold-start problem for standard VLMs.
We propose a training-free paradigm to empower VLMs with autonomous search capabilities via \textbf{cross-modal model merging}. By fusing a text-based search agent with a base VLM, we show that multi-modal search capabilities can be effectively composed without any additional multi-modal training data. To mitigate parameter interference during cross-modal integration, we introduce \textbf{Optimal Brain Merging (OBM)}, a saliency-aware merging algorithm that identifies task-critical parameters based on their impact on model loss using only a small set of calibration samples. 
Extensive experiments on search-intensive benchmarks (e.g., InfoSeek, MMSearch) reveal that: (1) Model merging secures a reasonable performance \textbf{floor} as a zero-shot agent, with OBM achieving superior search rates; (2) OBM significantly raises the performance \textbf{ceiling} as a warm-start strategy, achieving faster convergence and higher peak accuracy than standard VLM initialization.
\end{abstract}

\section{Introduction}

\begin{figure}
    \centering
    \includegraphics[width=\linewidth]{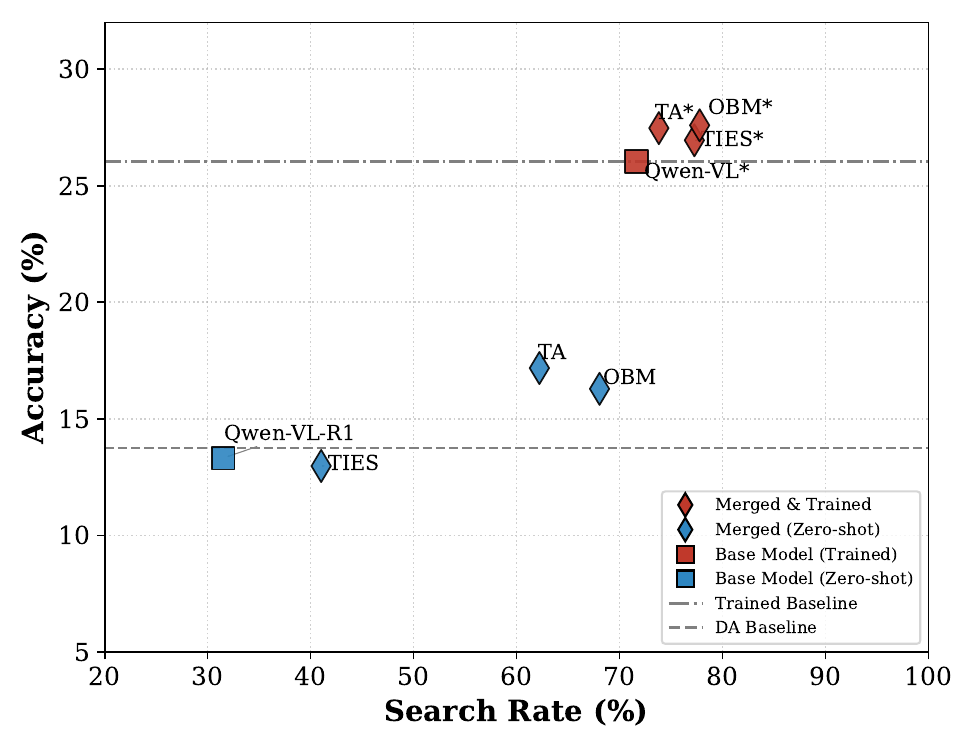}
    \caption{Performance of the merging-based paradigm versus standard VLM training. The DA Baseline denotes Direct Answering. Compared to the standard base VLM (squares), model merging (diamonds) establishes a higher performance floor in zero-shot settings (blue) and further raises the ceiling after reinforcement learning (red).}
    \vspace{-4mm}
    \label{fig:scatter}
\end{figure}

Recent advances in vision–language models (VLMs) have enabled systems capable of perceiving complex visual scenes and responding to diverse multi-modal queries. Beyond passive understanding, an emerging research direction focuses on building multi-modal search agents \citep{mmsearchr1}—models that can autonomously decide when external information is required, invoke appropriate tools (e.g., image or text search), and integrate retrieved evidence through multi-step reasoning. Such agents play a central role in real-world applications including open-domain visual question answering, visual fact verification, etc.
Despite their promise, constructing multi-modal search agents remains technically demanding. Existing approaches typically rely on large-scale supervised traces or reinforcement learning (RL) over carefully designed search environments. Collecting high-quality search-reasoning trajectories is costly, and RL training introduces substantial computational overhead and instability. More importantly, training often suffers from a pronounced cold-start problem: a vanilla VLM lacks intrinsic search behavior, forcing optimization to discover tool-use policies from scratch. These challenges significantly limit the accessibility and reproducibility of multi-modal agent research.

Interestingly, many of the capabilities required for effective multi-modal search agents are already present, but they are distributed across different types of models: vision–language models excel at visual perception and cross-modal grounding, while large language models demonstrate strong multi-step reasoning and tool-use abilities. The core challenge is therefore not acquiring these skills individually, but integrating them into a single system capable of both visual understanding and complex reasoning. A natural approach is parameter-level model fusion, which leverages pre-trained models of different modalities without requiring costly end-to-end multi-modal training. Existing merging methods, such as Task Arithmetic (TA) \citep{ta} or TIES-merging (TIES) \citep{ties}, provide a starting point for such fusion, but naively applying them may not fully capture the intricate dependencies between vision and language components. This motivates the need for fusion strategies that explicitly respect cross-modal alignment while combining complementary capabilities.

To address this limitation, we propose a cross-modal model merging framework that combines a vision–language model and a language-based search agent at the parameter level. Within this framework, we introduce Optimal Brain Merging (OBM), a saliency-aware parameter fusion algorithm tailored for cross-modal architectures. OBM estimates parameter importance based on their effect on model loss, using forward propagation on a small set of calibration samples to derive \textbf{activation-aware} saliency across the vision encoder, projector, and language backbone. By explicitly accounting for the structural dependencies between modalities, OBM enhances the robustness of the merged model and provides a principled way to adapt and extend existing merging techniques for effective cross-modal integration.
Building upon OBM, we construct multi-modal search agents by integrating a text-based search agent and a vision–language model starting from a shared pre-trained backbone. This parameter-level fusion enables autonomous tool use without any additional multi-modal training. The resulting agents establish a strong zero-shot performance floor and serve as a reliable warm-start for downstream reinforcement learning, allowing faster convergence and higher peak performance while preserving cross-modal consistency.

Extensive experiments on search-intensive benchmarks—including FVQA-test \citep{mmsearchr1}, InfoSeek \citep{infoseek}, MMSearch \citep{mmsearch}, and LiveVQA \citep{livevqa}—demonstrate that model merging yields complementary benefits. On one hand, merged agents exhibit substantially stronger search behavior in zero-shot settings, establishing a higher performance floor than prompt-based approaches. On the other hand, when used as initialization for RL, OBM consistently accelerates convergence and leads to higher peak performance, effectively raising the performance ceiling of multi-modal search agents.

In summary, this work makes the following contributions:
\begin{itemize}
    \item We present a new perspective on multi-modal search agent construction, reframing it as a problem of capability composition via model merging rather than end-to-end multi-modal training.
    \item We introduce OBM, a saliency-aware merging algorithm that leverages activation-dependent loss sensitivity to resolve parameter interference in cross-modal architectures.
    % \item We show that merging-based agents establish a reasonable zero-shot performance floor, exhibiting autonomous search behavior without requiring training.
    % \item We demonstrate that OBM serves as an effective warm-start strategy, significantly improving training efficiency and peak performance compared to standard VLM initialization.
    \item We show that merging-based agents not only establish a strong zero-shot performance floor, exhibiting autonomous search behavior without additional training, but also serve as an effective warm-start for downstream reinforcement learning, significantly improving training efficiency and peak performance compared to standard VLM initialization.
\end{itemize}

\begin{figure*}[t]
    \centering
    \includegraphics[width=0.85\textwidth]{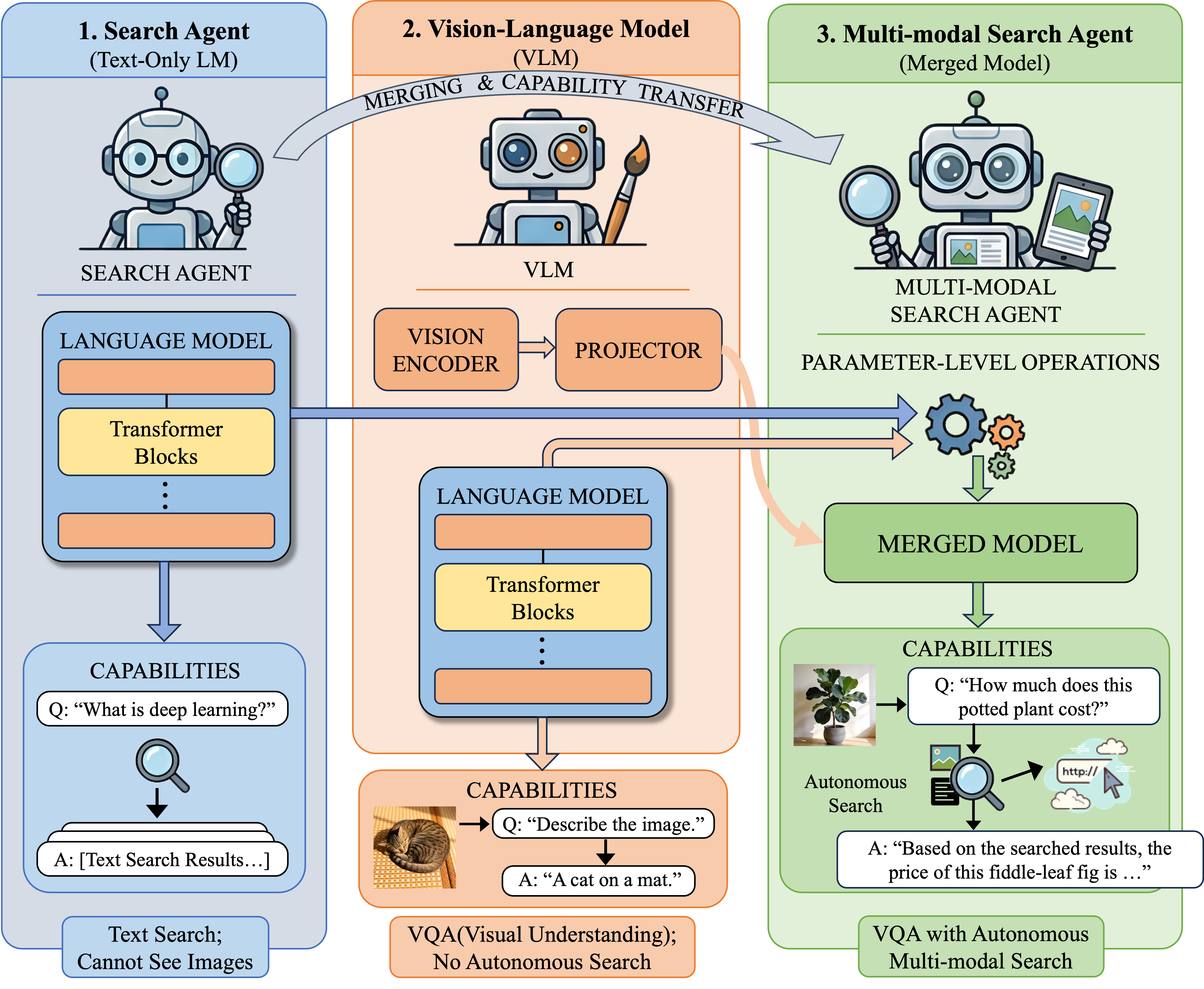}
    \caption{Overview of the merging paradigm for constructing a multi-modal search agent. The merged model retains the frozen vision encoder and projector from the VLM, while its language module is obtained via parameter-level merging between the VLM and the LLM. As a result, the merged agent is able to answer multi-modal queries by autonomously searching external information.}
    \label{fig:mmsa}
\end{figure*}

\section{Related Work}

\subsection{Model Merging in LLMs and VLMs}
Model merging has emerged as a powerful technique to combine specialized capabilities from multiple fine-tuned models without additional training. Early works like TA \citep{ta} demonstrate that subtracting pre-trained weights from fine-tuned ones creates "task vectors" that can be linearly composed. To address parameter interference and sign conflicts, TIES \citep{ties} introduces trimming and sign consensus, while DARE \citep{dare} utilizes random sparsification to eliminate redundant parameters. Although these methods excel in unimodal scenarios (e.g., merging multiple text-based LLMs), they rely solely on weight magnitudes or apply random pruning strategies, which may struggle to maintain the delicate alignment required in cross-modal architectures.
% \subsection{Cross-modal Model Merging}
Recent studies have begun exploring model merging for Vision-Language Models. \citet{viscodex} and \citet{bring} utilize TA to merge LLMs with VLMs to enhance reasoning in specific domain. However, these approaches are primarily evaluated on {single-turn reasoning tasks} and do not account for the complexities of autonomous tool-use. To the best of our knowledge, our work is the {first to explore model merging for constructing multi-modal search agents}, focusing on the seamless integration of visual perception and multi-step search-reasoning logic.

\subsection{Multi-modal Search Agents}
The development of agents capable of visual perception and autonomous tool-use typically relies on high-quality reasoning traces and extensive RL. MMSearch-R1 \citep{mmsearchr1} provides a representative framework in this domain, establishing a specialized pipeline and benchmark for fine-tuning VLMs to execute search-based reasoning. Parallel to this, large-scale systems such as WebWatcher \citep{webwatcher} explore the boundaries of general-purpose vision-language deep research agent by leveraging massive datasets and orchestrating a diverse array of tools (e.g., Code Interpreter, OCR).
In contrast, our work focuses specifically on the \textbf{compositional efficiency} of multi-modal search agents. Rather than seeking a universal tool-user through large-scale training, we investigate a novel \textbf{merging-based paradigm} that constructs a specialized search agent by fusing existing task-specific experts. This approach not only provides a training-free solution with a high performance floor but also serves as an optimal starting point for subsequent RL, offering a distinct and efficient research trajectory for developing advanced multi-modal search agents.

\subsection{Foundations of Saliency-aware Optimization}
Our proposed OBM algorithm is theoretically grounded in classic pruning and quantization literature. We draw inspiration from Optimal Brain Damage (OBD) \citep{obd}, which leverages second-order Taylor expansions (the Hessian matrix) to identify "salient" parameters that significantly impact the loss function. This saliency-aware perspective has been further refined in modern quantization methods like GPTQ \citep{gptq}, which utilizes activation-aware Hessian information to minimize reconstruction error. By adapting these insights to the model merging domain, OBM identifies task-critical parameters based on their actual contribution to cross-modal reasoning rather than simple numerical magnitude.

% \section{Preliminaries: Model Merging via Task Vectors}

\section{Methodology}
% In this section, we first delineate the architectural framework for composing a multi-modal search agent through model merging. Subsequently, we introduce OBM, a saliency-aware algorithm designed to resolve parameter interference and enhance cross-modal capability transfer.

In this section, we first introduce the overall framework for composing a multi-modal search agent via model merging, building upon standard formulations of task vectors and existing merging paradigms. We then present OBM, a saliency-aware algorithm designed to resolve parameter interference and facilitate effective cross-modal capability transfer.

\subsection{Preliminaries: Model Merging via Task Vectors}
Given a pre-trained base model $\theta_B$ and a fine-tuned checkpoint $\theta_T$ for a specific task $T$, the \textit{task vector} is defined as $\delta_T = \theta_T - \theta_B$. The goal of model merging is to combine multiple task vectors $\{\delta_1, \delta_2, \dots, \delta_n\}$ into a single unified model $\theta_{merged}$. 

A common baseline is \textbf{TA} \citep{ta}, which performs a linear summation: $\theta_{merged} = \theta_B + \sum \lambda_i \delta_i$. To address parameter interference, methods like \textbf{TIES} \citep{ties} introduce a two-stage process: (1) \textit{Trimming}, which keeps only the top-$k\%$ values based on magnitude $|\delta_i|$; and (2) \textit{Sign Consensus}, which resolves direction conflicts via a majority vote. However, these methods rely purely on the weight values in isolation, ignoring the underlying data distribution and inter-module dependencies.

\subsection{Model Merging for Multi-modal Search Agent}
Our goal is to create an agent that possesses both the visual perception of a VLM and the autonomous tool-use reasoning of a search agent. We leverage the fact that both Search-R1 \citep{searchr1} and the language backbone of \texttt{Qwen2.5-VL-7B-Instruct} \citep{qwenvl} share the same ancestry, having been fine-tuned from the \texttt{Qwen2.5-7B} \citep{qwen2025qwen25technicalreport} base model.
Let $\theta_{B} \in \mathbb{R}^d$ denotes the parameters of the shared base model. We define the task vectors for the text-based search agent ($\delta_{S}$) and the language module of the VLM ($\delta_{V}$) as follows:
\begin{equation}
    \delta_{S} = \theta_{Search} - \theta_{B}, \quad \delta_{V} = \theta_{VL} - \theta_{B},
\end{equation}
where $\theta_{Search}$ and $\theta_{VL}$ represent the weights of Search-R1 \footnote{Search-R1: \path{SearchR1-nq_hotpotqa_train-qwen2.5-7b-em-ppo-v0.3}.} and the Qwen-VL \footnote{Qwen-VL: \path{Qwen2.5-VL-7B-Instruct}.} 's language module, respectively.

To construct the multi-modal agent, we transplant the vision encoder and the projector from the VLM directly into the fused system. The language module weights are initially combined using Task Arithmetic:
\begin{equation}
    \theta_{merged} = \theta_{B} + \lambda_{S} \delta_{S} + \lambda_{V} \delta_{V},
\end{equation}
where $\lambda$ are scaling factors.
This architecture ensures the model can process visual inputs while inheriting the tool-use reasoning logic. 

\subsection{Optimal Brain Merging (OBM)}
To overcome the \textbf{modality-agnostic} limitations discussed in the Introduction and more accurately identify parameters critical for cross-modal reasoning, we propose OBM. OBM introduces a saliency-aware framework consisting of two stages: saliency-driven sparsification and saliency-weighted sign consensus.

\textbf{Stage 1: Saliency-driven Sparsification.} The core principle of OBM is to prune task vector entries that have a negligible impact on the model's loss $\mathcal{L}$. Drawing inspiration from OBD\citep{obd}, we quantify the \textit{saliency} $s_i$ of the $i$-th component of a task vector $\delta_i$ by the second-order Taylor expansion of the loss change:
\begin{equation}s_i = \frac{1}{2} \frac{\partial^2 \mathcal{L}}{\partial \theta_i^2} \delta_i^2 = \frac{1}{2} h_{ii} \delta_i^2,
\end{equation}
where $h_{ii}$ represents the $i$-th diagonal element of the Hessian matrix.
To avoid the prohibitive computational cost of calculating the full Hessian for large-scale models, we adopt a layer-wise Mean Squared Error (MSE) approximation, as motivated by GPTQ\citep{gptq}. For the $l$-th linear layer with weight matrix $\mathbf{W}^l$ and input activations $\mathbf{X}^l$, the local objective is defined as 
\begin{equation}
    \Delta \mathcal{L}^l = \| \Delta \mathbf{W}^l \mathbf{X}^l \|_2^2.
\end{equation}
Under this formulation, the Hessian matrix simplifies to $\mathbf{H}^l = 2\mathbf{X}^l {\mathbf{X}^l}^{\top}$. This enables efficient saliency computation via a single forward pass using a minimal set of calibration samples (multi-modal VQA data for $\delta_{V}$ and search-reasoning data for $\delta_{S}$). Crucially, since $\mathbf{X}^l$ is dynamically collected during the forward pass, the saliency of $\delta_V$ inherently captures the activation-aware dependencies between the language backbone and the frozen vision-projection modules. This effectively bridges the gap between disparate modalities, whereas the saliency of $\delta_S$ isolates the logic governing search-based reasoning. For each task vector, we retain only the top-$p\%$ parameters with the highest saliency to eliminate interference while preserving modality-specific expertise:
\begin{equation}
\hat{\delta}_i =
\begin{cases}\delta_i & \text{if } s_i \in \text{Top-}p\%  \\ 
0 & \text{otherwise.}
\end{cases}
\end{equation}
For non-linear components such as embedding layers, we assign uniform saliency to all parameters, as task-specific knowledge is primarily concentrated within the Transformer blocks. This allows these layers to follow the same subsequent merging pipeline while maintaining consistent model density.

\textbf{Stage 2: Saliency-weighted Sign Consensus.} In cases where task vectors exhibit conflicting directions (i.e., $\text{sign}(\delta_{S,i}) \neq \text{sign}(\delta_{V,i})$), traditional methods \citep{ties, della} rely on simple magnitude comparison. We instead propose a \textbf{weighted consensus mechanism} that accounts for the relative importance of each task. The consensus direction is determined by:
\begin{equation}
\sigma_i = \text{sign} \left( \sum_{k \in {S, V}} s_{i}^{(k)} \cdot \text{sign}(\hat{\delta}{i}^{(k)}) \right).
\end{equation}
The final merged task vector is construc    ted by aligning the parameters with this consensus sign. Specifically, we preserve the values that agree with $\sigma_i$, ensuring that parameters vital to "multi-modal synergy" are retained with their correct task-specific orientations.

\section{Experiments}
We conducted experiments in two phases, before and after training, to explore the superiority of the model merging paradigm in constructing multi-modal search agent.

\subsection{Experimental Settings}
We evaluate the effectiveness of our merging-based paradigm across various VQA benchmarks. We compare our proposed OBM with standard VLM baselines and existing model merging algorithms.

\paragraph{Baselines and Answering Paradigms}
To rigorously assess the "floor" and "ceiling" effects of our paradigm, we consider several answering workflows and baselines. All models (including merged variants and the trained baseline) utilize a consistent search toolset comprising both image-to-image and text-to-text search engines. Please refer to Appendix ~\ref{app:search_tool} for implementation details of search tools.
\begin{itemize}
    \item \textbf{Direct Answering (DA)}: A standard VLM baseline where Qwen-VL generates an answer directly based on the input image and question without any external search.
    \item \textbf{Search-augmented (Qwen-VL-R1)}: We prompt the vanilla Qwen-VL to act as an agent. It is instructed to call search tools only when the visual information is insufficient. This serves as a zero-shot prompting baseline.
    \item \textbf{RAG-based Workflow}: Following \citep{mmsearchr1}, this fixed multi-step pipeline performs image search, generates a text query, conducts text search, and synthesizes the final answer from all retrieved information.
    % Following the pipeline in \citep{mmsearchr1}, this is a fixed multi-step process: (1) An image search is performed; (2) The model generates a text query based on the image search results; (3) A text search is conducted; (4) The final answer is synthesized based on the given image and question, and multiple modality searched results. This represents a strong non-autonomous baseline.
    \item \textbf{MMSearch-R1 (Trained Baseline)}: An RL-trained baseline fine-tuned from Qwen-VL on multi-modal data with both search-required and search-free examples. Our reported results use a reproduced version due to tool unavailability.
    % A critical baseline representing the "RL-trained" approach. It is fine-tuned from Qwen-VL using high-quality multi-modal data including both search-required and search-free VQA examples. We reproduced this model using the official training data and our search toolset; because search tools are not open source and consistency cannot be guaranteed, and in order to observe the training trends of the model more comprehensively, all reported results for MMSearch-R1 are based on our reproduced version.
    \item \textbf{Merging-based Agents (TA, TIES, DARE, OBM)}: These models constructed by merging a text-based search agent (Search-R1) with Qwen-VL. For OBM, we computed the parametric saliency of Search-R1 using the NQ-HotpotQA dataset \citep{nq, hotpotqa} merged by \citep{searchr1}, and that of the language modules in Qwen-VL using MMStar \citep{mmstar}, with 128 randomly sampled questions per dataset. See Appendix \ref{app:merging_hyperparameters} for merging hyper-parameters.
    % These models are constructed by merging a text-based search agent (Search-R1) and the Qwen-VL. Like MMSearch-R1, these agents autonomously decide when and how to use search tools. For OBM, we computed the parametric saliency of Search-R1 using the NQ-HotpotQA dataset \citep{nq, hotpotqa} merged by \citep{searchr1}, and that of the language modules in Qwen-VL using MMStar \citep{mmstar}. In both cases, we randomly sampled 128 examples—questions alone, without requiring ground-truth answers—from each dataset. See the Appendix for detailed hyper-parameter configurations.
\end{itemize}
We have largely reused the prompts from \citep{mmsearchr1}, with minor tweaks. See the Appendix \ref{app:prompts} for details.

\paragraph{Benchmarks}
We evaluate models on four datasets, including both knowledge-intensive and information-seeking VQA tasks, to assess their ability to autonomously decide whether to use a search tool to assist in answering questions. We explicitly excluded SimpleVQA \citep{simplevqa} used in \citep{mmsearchr1}, because our preliminary experiments Appendix ~\ref{app:simplevqa} indicated its extremely low dependency on external search (where even the trained MMSearch-R1 and Qwen-VL with RAG workflow underperform compared to direct VLM answering). The chosen benchmarks are:
\begin{itemize}
    % \item \textbf{FVQA-test}: A manually curated set of 1,800 examples derived from FVQA-auto-vc, the InfoSeek Human Split (re-annotated), and new human-collected samples. It covers diverse visual and textual knowledge categories.
    % \item \textbf{InfoSeek}: A large-scale dataset focusing on information-seeking VQA grounded in Wikidata triples. We use the version of MMSearch-R1 published, which randomly sampled 2,000 examples from its test split to ensure evaluation efficiency.
    % \item \textbf{MMSearch}: A benchmark designed for real-world search tasks, featuring subdomains like News (up-to-date events from 2024) and rare Knowledge. Following MMSearch-R1, we use the 171 visual question-answer pairs which require multi-modal search to solve.
    % \item \textbf{LiveVQA}: A timely dataset built from global news platforms (e.g., CNN, BBC). It contains complex, multi-hop questions that require reasoning over both visual content and up-to-date textual information.
    \item \textbf{FVQA-test}: A curated set of examples from FVQA-auto-vc, re-annotated InfoSeek, and additional human-collected samples, covering diverse visual and textual knowledge.
    \item \textbf{InfoSeek}: A subset of the published MMSearch-R1 test split, focusing on information-seeking VQA grounded in Wikidata triples.
    \item \textbf{MMSearch}: Examples from subdomains such as News (2024 events) and rare Knowledge, all requiring multi-modal search.
    \item \textbf{LiveVQA}: Multi-hop, timely questions collected from global news platforms (e.g., CNN, BBC) requiring reasoning over both visual and textual content.
\end{itemize}

\paragraph{Metrics}
To comprehensively evaluate the performance and autonomous behavior of the multi-modal search agents, we employ the following metrics:
% \begin{itemize}
% \item \textbf{Accuracy ($Acc$):} We use the Exact Match (EM) score between the model's prediction and the ground-truth answer. Following standard VQA evaluation protocols, both the prediction and the gold answer undergo a normalization process before comparison, which includes converting to lowercase, removing punctuation, stripping articles (e.g., "a", "an", "the"), and eliminating redundant whitespaces.
% \item \textbf{Search Rate ($SR$):} This metric quantifies the frequency with which the agent autonomously decides to invoke external knowledge. $SR$ is defined as the percentage of test cases where the model's reasoning trajectory contains at least one call to either the image-to-image or text-to-text search tool.
% \item \textbf{Image and Text Search Rates ($SR_{img}$, $SR_{txt}$):} To provide a more granular view of tool-use preferences, we separately report the invocation rates for the image search engine ($SR_{img}$) and the text search engine ($SR_{txt}$). A specific query is counted toward these metrics if the corresponding tool appears at least once in its reasoning path.
% \end{itemize}
\begin{itemize}
    \item \textbf{Accuracy ($Acc$):} Exact Match (EM) between the model prediction and ground-truth answer, following standard VQA normalization (lowercasing, punctuation removal, stripping articles, and trimming whitespace).
    \item \textbf{Search Rate ($SR$):} The percentage of test cases where the agent invokes any external search tool (image or text) at least once.
    \item \textbf{Image and Text Search Rates ($SR_{img}$, $SR_{txt}$):} Separate invocation rates for image and text search tools, counted if the tool appears at least once in the reasoning trajectory.
\end{itemize}
% Note that $SR_{img}$ and $SR_{txt}$ are not mutually exclusive; multi-hop reasoning can result in $SR_{img} + SR_{txt} > 100\%$. Additional metrics such as hybrid search rate are reported in the Appendix.
It is important to note that $SR_{img}$ and $SR_{txt}$ are not mutually exclusive. A sophisticated agent may perform a multi-hop search (e.g., first identifying an entity via image search and then retrieving specific facts via text search) within a single episode. Consequently, while $SR$ represents the union of search events, the sum of $SR_{img}$ and $SR_{txt}$ may exceed $100\%$, reflecting the agent's multi-step reasoning depth.

\begin{table*}[t]
\caption{Benchmark results of our merged models vs. baselines. We compare our merged models—derived via TA, TIES, DARE, and OBM—against several baselines, including the trained MMSearch-R1 and three Qwen-VL variants: Direct Answering (DA), Search-augmented (R1), and RAG-based workflows.}
\label{tab:zero-shot}
\resizebox{\textwidth}{!}{%
\begin{tabular}{c|ccc|ccc|ccc|ccc}
\hline
\multirow{2}{*}{\textbf{Model}} & \multicolumn{3}{c|}{\textbf{FVQA-test}} & \multicolumn{3}{c|}{\textbf{InfoSeek}} & \multicolumn{3}{c|}{\textbf{MMSearch}} & \multicolumn{3}{c}{\textbf{LiveVQA}} \\
 & \textbf{Acc.} & \textbf{SR\_img} & \textbf{SR\_txt} & \textbf{Acc.} & \textbf{SR\_img} & \textbf{SR\_txt} & \textbf{Acc.} & \textbf{SR\_img} & \textbf{SR\_txt} & \textbf{Acc.} & \textbf{SR\_img} & \textbf{SR\_txt} \\ \hline
MMSearch-R1 & 29.39 & 66.06 & 47.28 & 34.60 & 56.20 & 36.80 & 24.56 & 86.55 & 67.25 & 15.60 & 77.67 & 53.16 \\ \hline
Qwen-VL-DA & 15.94 & 0.00 & 0.00 & 17.90 & 0.00 & 0.00 & 8.77 & 0.00 & 0.00 & 12.41 & 0.00 & 0.00 \\
Qwen-VL-R1 & 15.61 & 8.44 & 25.50 & 19.45 & 11.85 & 16.50 & 10.53 & 16.38 & 45.03 & 7.69 & 12.54 & 22.43 \\
Qwen-VL-RAG & 24.83 & 100.00 & 100.00 & 21.30 & 100.00 & 100.00 & 21.64 & 100.00 & 100.00 & 13.69 & 100.00 & 100.00 \\ \hline
TA & 21.50 & 19.61 & 43.11 & 22.85 & 35.55 & 38.80 & 13.45 & 15.21 & 64.92 & 10.91 & 38.01 & 43.20 \\
TIES & 15.00 & 1.94 & 32.72 & 15.10 & 2.45 & 37.00 & 9.94 & 4.09 & 49.12 & 11.88 & 6.47 & 37.70 \\
DARE & 10.06 & 8.39 & 29.94 & 10.65 & 12.50 & 35.75 & 4.09 & 5.26 & 32.16 & 5.47 & 13.63 & 17.71 \\
OBM & 20.67 & 16.50 & 55.55 & 19.65 & 31.70 & 46.15 & 13.45 & 21.63 & 78.94 & 11.38 & 19.10 & 60.05 \\ \hline
\end{tabular}%
}
\end{table*}

\subsection{Zero-shot Performance: Securing a High Floor}
We first evaluate the performance of the training-free merging paradigm in Table \ref{tab:zero-shot}. The results provide several key insights into how model merging establishes a superior "floor" for multi-modal search agents.

\paragraph{Limitations of Prompt-based Autonomy} Simply relying on sophisticated prompting (\textbf{Qwen-VL-R1}) to elicit autonomous tool-use yields suboptimal results. While Qwen-VL-R1 shows marginal improvements over Direct Answering (\textbf{Qwen-VL-DA}) on information-seeking benchmarks—specifically on InfoSeek (19.45\% vs. 17.90\%) and MMSearch (10.53\% vs. 8.77\%)—its overall performance remains constrained. This suggests that the inherent reasoning logic for visual search is not fully activated by prompts alone when the base model lacks dedicated search-reasoning parameters.

\paragraph{Merging Transfers Search "Instincts"} In contrast, merging-based agents (\textbf{TA} and \textbf{OBM}) significantly outperform Qwen-VL-R1 across all benchmarks. Notably, on FVQA-test and MMSearch, they achieve an accuracy gain of approximately 5\% over Qwen-VL-DA. By examining the search behaviors, we observe that these models exhibit remarkably high search rates. For instance, \textbf{OBM achieves a text search rate ($SR_{txt}$) of 78.94\% on MMSearch}. This high activation demonstrates that the integration of Search-R1 parameters successfully transposes search logic into the VLM, enabling the model to autonomously seek external evidence to resolve visual queries.

\paragraph{Narrowing the Gap to RAG} Although TA and OBM do not yet consistently surpass the Qwen-VL-RAG baseline in a zero-shot setting, they significantly narrow the performance gap compared to the Qwen-VL-R1. It is important to note that the RAG baseline follows a \textbf{rigid, predefined workflow} (sequential image and text search) which, while robust, is not necessarily optimal for every query. In contrast, our merging-based agents attempt to replicate the \textbf{autonomous reasoning} seen in trained models like MMSearch-R1. The fact that merged models can relatively approach RAG-level accuracy without a hard-coded pipeline demonstrates their potential. This suggests that model merging provides a much sturdier "floor" for building flexible agents that can eventually surpass fixed workflows through more efficient, targeted tool-use.

\paragraph{Modality Preference and Search Efficiency} An analysis of the $SR_{img}$ and $SR_{txt}$ distribution reveals a distinctive "text-centric" bias in zero-shot merged models. While the RL-trained \textbf{MMSearch-R1} demonstrates high efficiency by achieving superior accuracy with a more balanced search frequency, zero-shot merged models rely more heavily on textual reasoning. This indicates that while merging successfully implants the "will" to search, the \textbf{fine-grained coordination} between visual evidence and specific tool selection—such as knowing exactly when to trigger an image search versus a text search—is a higher-order skill that requires further optimization.

\paragraph{Vulnerability of Unimodal Merging in Cross-modal Contexts} Standard algorithms for unimodal model merging, such as \textbf{TIES} \citep{ties} and \textbf{DARE} \citep{dare}, struggle in this cross-modal scenario. Notably, \textbf{DARE} suffers from a catastrophic performance drop compared to the base Qwen-VL-DA. We attribute this to DARE’s mechanism of randomly dropping task vector entries. In a VLM, the language module parameters are delicately aligned with the vision encoder and projector; DARE's stochastic pruning likely disrupts these cross-modal synergistic dependencies, effectively severing the pathway for visual information. Similarly, magnitude-based methods like \textbf{TIES} may inadvertently eliminate small-magnitude parameters that are nonetheless critical for bridging visual perception and search reasoning. These failures reinforce the necessity of \textbf{OBM}, which utilizes saliency-aware Hessian information to preserve task-critical parameters while maintaining modality alignment.

Based on these zero-shot observations, we select \textbf{TA}, \textbf{TIES}, and \textbf{OBM} as the base models for the subsequent training phase. While TIES performs slightly worse than TA, it maintains a reasonable search rate and serves as a representative baseline for interference-reduction methods. \textbf{OBM}, despite having slightly lower accuracy than TA on some benchmarks, exhibits the highest overall search rate. We hypothesize that this stronger "search-reasoning instinct" indicates a higher performance ceiling when the model is further optimized through reinforcement learning, which we explore in the next section.

\subsection{Training Dynamics: Raising the Performance Ceiling}
\begin{figure}
    \centering
    \includegraphics[width=\linewidth]{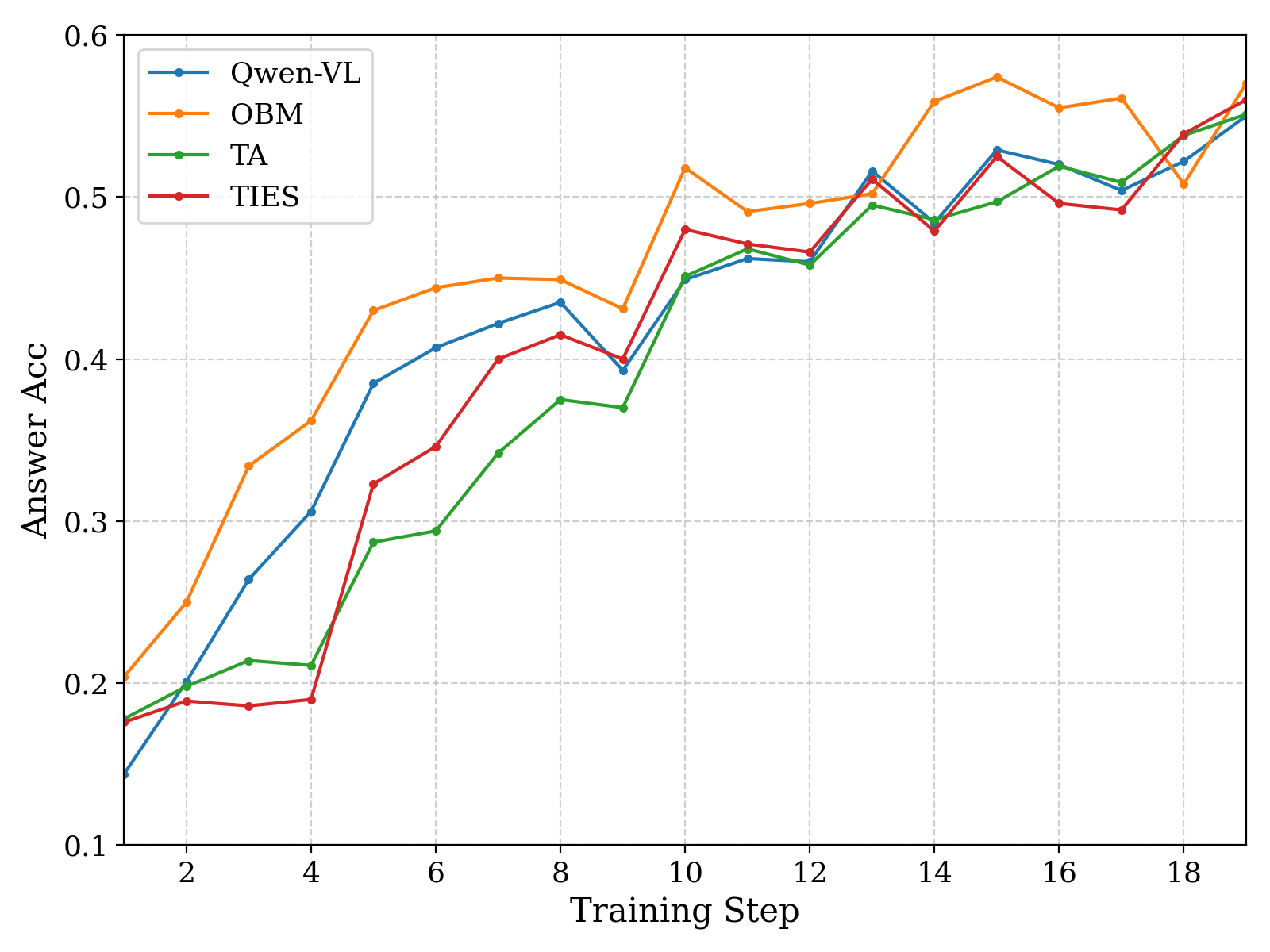}
    \caption{Comparison of training accuracy in the early training phase across different base models.}
    \label{fig:early_stage}
\end{figure}

To evaluate the impact of model merging on the training process, we performed reinforcement learning using the same training data (5,000 samples) and parameter update frequency as in MMSearch-R1~\citep{mmsearchr1}. All other training hyper-parameters were also kept identical to those used in MMSearch-R1.

\paragraph{Accelerated Convergence}
The initial phase of training reveals a significant efficiency gap between standard VLM initialization and our OBM based paradigm. As shown in Figure \ref{fig:early_stage}, which tracks the training accuracy over the first two epochs (18 steps), OBM continuously and substantially outperforms all other initialization strategies. Notably, OBM requires only 5 steps to surpass the 1-epoch performance of other models and reaches their 2-epoch accuracy level within just 14 steps. This rapid ascent indicates that the saliency-aware merging in OBM preserves a pre-aligned reasoning-perception space, allowing the model to bypass the "cold-start" phase typical of training multi-modal agents from scratch.

\begin{figure}
    \centering
    \includegraphics[width=\linewidth]{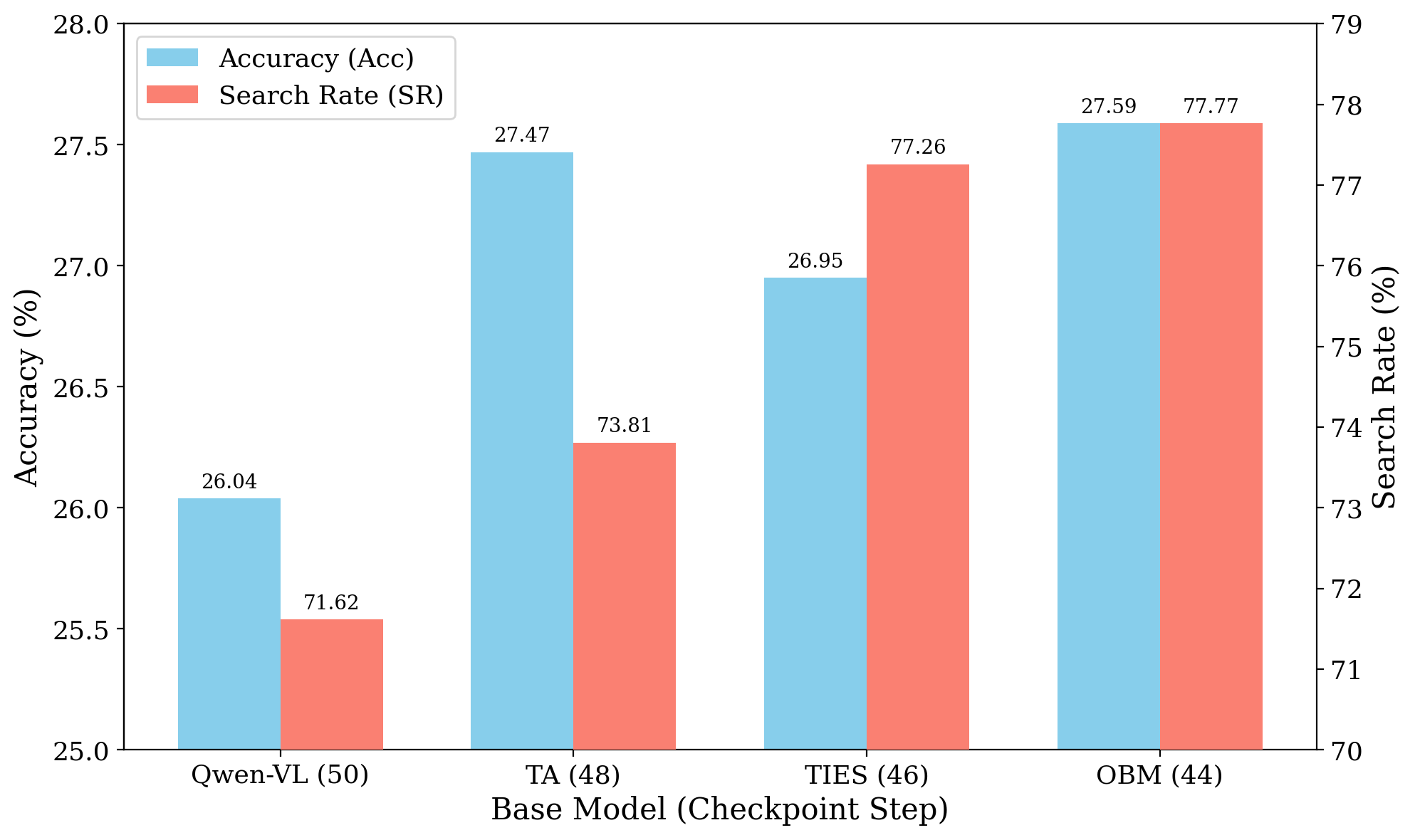}
    \caption{Average accuracy and search rate on benchmarks at each model’s best checkpoint.}
    \label{fig:optimal}
\end{figure}

\paragraph{Breakthrough in Peak Performance}
A central claim of our work is that merging "raises the ceiling" of the agent's ultimate capability. Figure \ref{fig:optimal} summarizes the average $Acc$ and $SR$ at each model’s optimal checkpoint. The results clearly demonstrate that starting from vanilla Qwen-VL leads to a lower performance plateau compared to starting from a merged model. OBM achieves the highest peak performance (27.59\% Avg. Acc and 77.77\% Avg. SR) at step 44, which is not only better but also earlier than the peak of other variants. This suggests that the "search instincts" embedded during the merging phase provide a more robust foundation for the model to explore and master complex tool-use logic.

\paragraph{Generalization vs. In-domain Fitting}
We further analyze the late-stage training performance (steps 44–52) in Table \ref{tab:trained_acc}. Several key observations emerge regarding the stability and generalization of the merged agents:
\begin{itemize}
    \item \textbf{Superiority of Merged Initializations}: Across the entire late-stage window, the pure Qwen-VL baseline (\textbf{MMSearch-R1}) consistently lags behind the merged models in both in-domain (FVQA-test, InfoSeek) and out-of-distribution (MMSearch, LiveVQA) benchmarks. This confirms that the benefits of model merging persist throughout the training lifecycle.
    \item \textbf{The Overfitting Risk of TIES}: While \textbf{TIES} shows impressive fitting capabilities on in-domain data (leading on FVQA-test and InfoSeek), it suffers from poor generalization. Its performance on the search-intensive MMSearch benchmark is notably weaker than TA and OBM, suggesting that magnitude-based merging may lead to a more rigid parameter distribution that overfits to specific reasoning patterns.
    \item \textbf{Robust Generalization of OBM}: Both \textbf{TA} and \textbf{OBM} demonstrate superior generalization on OOD benchmarks like MMSearch and LiveVQA. OBM, in particular, maintains a high "performance-to-step" ratio, hitting the global SOTA at step 44 and continuing to exhibit strong results at step 52.
\end{itemize}

In summary, the training-free OBM paradigm serves as a powerful warm-start strategy. By precisely integrating task-specific expertise while mitigating interference, OBM enables the agent to converge faster and ultimately reach a performance ceiling that is unattainable through standard multi-modal training alone.

\begin{table}[]
\caption{Late-stage performance comparison of various base models. We report the benchmark results during the final training phase (steps 44–52).}
\label{tab:trained_acc}
\resizebox{\columnwidth}{!}{%
\begin{tabular}{c|c|cccc|c}
\hline
\textbf{Step} & \textbf{Model} & \textbf{FVQA-test} & \textbf{InfoSeek} & \textbf{MMSearch} & \textbf{LiveVQA} & \textbf{Avg.} \\ \hline
\multirow{4}{*}{44} & Qwen-VL & 27.72 & 33.45 & 23.98 & 15.24 & 25.10 \\
 & TA & 29.06 & 34.25 & 26.90 & 15.55 & 26.44 \\
 & TIES & \textbf{30.28} & \textbf{34.85} & 25.73 & 15.71 & 26.64 \\
 & OBM & 28.94 & 34.45 & \textbf{30.99} & \textbf{15.96} & \textbf{27.59} \\ \hline
\multirow{4}{*}{46} & Qwen-VL & 28.39 & 33.80 & 22.22 & 15.21 & 24.91 \\
 & TA & 29.94 & 34.45 & \textbf{27.48} & 15.46 & 26.83 \\
 & TIES & \textbf{30.22} & \textbf{35.60} & 26.32 & 15.66 & \textbf{26.95} \\
 & OBM & 28.56 & 34.60 & 26.32 & \textbf{16.05} & 26.38 \\ \hline
\multirow{4}{*}{48} & Qwen-VL & 28.44 & 34.25 & 23.98 & 15.24 & 25.48 \\
 & TA & 28.61 & 34.75 & \textbf{30.41} & \textbf{16.13} & \textbf{27.47} \\
 & TIES & \textbf{30.17} & \textbf{35.75} & 23.98 & 15.94 & 26.46 \\
 & OBM & 29.00 & 34.60 & 25.15 & 15.99 & 26.19 \\ \hline
\multirow{4}{*}{50} & Qwen-VL & 29.39 & 34.60 & 24.56 & 15.60 & 26.04 \\
 & TA & 29.00 & 34.15 & \textbf{28.66} & 15.63 & \textbf{26.86} \\
 & TIES & \textbf{30.17} & 34.80 & 22.22 & \textbf{15.91} & 25.78 \\
 & OBM & 29.44 & \textbf{34.95} & 26.32 & 15.74 & 26.61 \\ \hline
\multirow{4}{*}{52} & Qwen-VL & 29.05 & 34.40 & 24.56 & 15.32 & 25.83 \\
 & TA & 29.06 & 34.85 & 28.07 & 15.91 & 26.97 \\
 & TIES & \textbf{30.06} & 34.90 & 20.47 & 15.77 & 25.30 \\
 & OBM & 28.89 & \textbf{34.95} & \textbf{29.83} & \textbf{15.96} & \textbf{27.41} \\ \hline
\end{tabular}%
}
\end{table}

\subsection{Ablation Study}

\begin{table}[]
\caption{Ablation study on calibration data for OBM. We compare the default OBM with OBM (Text-only) (text-based calibration) and Oracle (benchmark-specific calibration).}
\label{tab:ablation}
\resizebox{\columnwidth}{!}{%
\begin{tabular}{l|ccc|ccc}
\hline
\multirow{2}{*}{\textbf{Variant}} & \multicolumn{3}{c|}{\textbf{FVQA-test \& InfoSeek}} & \multicolumn{3}{c}{\textbf{MMSearch   \& LiveVQA}} \\
 & \textbf{Acc} & \textbf{SR\_img} & \textbf{SR\_txt} & \textbf{Acc} & \textbf{SR\_img} & \textbf{SR\_txt} \\ \hline
\textbf{OBM} & \textbf{20.16} & 24.10 & 50.85 & \textbf{12.42} & 20.37 & \textbf{69.50} \\
OBM (Text-only) & 19.48 & 24.21 & \textbf{53.29} & 12.20 & 23.96 & 65.53 \\
Oracle & 19.84 & \textbf{30.40} & 52.85 & 11.38 & \textbf{27.98} & 66.74 \\ \hline
\end{tabular}%
}
\end{table}

To analyze the impact of calibration data used for saliency computation of Qwen-VL in OBM, we compare the default OBM against two variants: \textbf{OBM (Text-only)}—calibrated solely on textual QA data—and \textbf{Oracle}—calibrated on samples drawn from the evaluation benchmarks. All variants use 128 calibration samples.

As shown in Table ~\ref{tab:ablation}, the default OBM consistently outperforms OBM (Text-only). Without visual forward passes during calibration, the Hessian fails to capture activation-aware dependencies between the language backbone and the visual projector, confirming that multi-modal calibration is essential to preserve cross-modal alignment during parameter integration.

Interestingly, despite being distribution-aligned, the Oracle variant yields lower accuracy than OBM. While it significantly boosts  $ SR_{\text{img}} $ , this leads to an "excessive but inefficient" image search behavior at the zero-shot stage: the agent triggers tools more frequently but cannot effectively leverage the surge of image retrieval. Overfitting to task-specific signals disrupts the delicate perception–reasoning balance.

Nonetheless, performance differences among the three variants are minor, underscoring OBM’s robustness—it thrives on task-agnostic, modality-aware calibration without requiring benchmark-specific cherry-picking.

\subsection{Merging Cost}

\begin{table}[t]
\centering
\caption{Computational cost of different merging methods. 
For OBM, saliency computation is performed once using forward propagation and can be cached for reuse.}
\label{tab:merging_cost}
\resizebox{\columnwidth}{!}{%
\begin{tabular}{lcccc}
\toprule
Method 
& \#Models 
& GPU 
& Saliency Computation 
& Merging Time \\
\midrule
TA   & 2 & 1$\times$H20 (96G) & --     & 0'22 \\
TIES & 2 & 1$\times$H20 (96G) & --     & 7'03 \\
OBM  & 2 & 1$\times$H20 (96G) & 2'31   & 6'46 \\
\bottomrule
\end{tabular}%
}
\end{table}

Table ~\ref{tab:merging_cost} reports the computational cost of different merging methods.  
All approaches complete model merging in just seconds to a few minutes on a single GPU.
OBM introduces a modest additional cost of approximately 2 minutes for saliency estimation, performed once via forward propagation on a small calibration set.  
This step requires no backpropagation, gradient updates, or iterative optimization, and the computed saliency can be cached and reused for all subsequent merges.

\section{Conclusion}
In this work, we revisit multi-modal search agents and propose a training-free, cross-modal model fusion paradigm. By integrating pre-trained vision–language models with language-based search agents and using OBM to preserve cross-modal dependencies, our approach enables autonomous tool use without additional multi-modal training. Experiments show that the resulting agents achieve strong zero-shot performance and provide an effective warm-start for reinforcement learning, demonstrating that parameter-level fusion is a practical alternative to end-to-end multi-modal training.

\newpage
\section*{Impact Statement}
This work aims to advance the field of artificial intelligence by introducing a training-free paradigm for constructing multimodal search agents through cross-modal model merging. The proposed techniques prioritize model efficiency, accessibility, and reproducibility, and are intended for general-purpose research use. While our approach may be applied to downstream tasks involving multimodal information retrieval and reasoning over visual and textual data, it neither introduces new data sources nor fundamentally extends the capabilities of existing vision–language models. Consequently, we do not anticipate significant negative ethical or societal impacts beyond those already associated with large-scale vision–language systems and automated information-seeking agents.
% Authors are \textbf{required} to include a statement of the potential broader
% impact of their work, including its ethical aspects and future societal
% consequences. This statement should be in an unnumbered section at the end of
% the paper (co-located with Acknowledgements -- the two may appear in either
% order, but both must be before References), and does not count toward the paper
% page limit. In many cases, where the ethical impacts and expected societal
% implications are those that are well established when advancing the field of
% Machine Learning, substantial discussion is not required, and a simple
% statement such as the following will suffice:

% ``This paper presents work whose goal is to advance the field of Machine
% Learning. There are many potential societal consequences of our work, none
% which we feel must be specifically highlighted here.''

% The above statement can be used verbatim in such cases, but we encourage
% authors to think about whether there is content which does warrant further
% discussion, as this statement will be apparent if the paper is later flagged
% for ethics review.


\bibliography{example_paper}
\bibliographystyle{icml2026}

%%%%%%%%%%%%%%%%%%%%%%%%%%%%%%%%%%%%%%%%%%%%%%%%%%%%%%%%%%%%%%%%%%%%%%%%%%%%%%%
%%%%%%%%%%%%%%%%%%%%%%%%%%%%%%%%%%%%%%%%%%%%%%%%%%%%%%%%%%%%%%%%%%%%%%%%%%%%%%%
% APPENDIX
%%%%%%%%%%%%%%%%%%%%%%%%%%%%%%%%%%%%%%%%%%%%%%%%%%%%%%%%%%%%%%%%%%%%%%%%%%%%%%%
%%%%%%%%%%%%%%%%%%%%%%%%%%%%%%%%%%%%%%%%%%%%%%%%%%%%%%%%%%%%%%%%%%%%%%%%%%%%%%%
\newpage
\appendix
\onecolumn

\section{Prompts}
\label{app:prompts}
The prompts used to reproduce MMSearch-R1 \citep{mmsearchr1} are exactly the same as those they released open-source:

\begin{promptbox}{First Round User Prompt}
\small
Answer the user's question based on the provided image. Examine the image carefully and identify any recognizable entities, such as faces, objects, locations, events, logos, or text. Determine whether you have sufficient knowledge to confidently recognize the main visual element and answer the user's question. If so, first explain your reasoning, then provide a clear and direct answer. \\
If you are unable to confidently identify the visual element, stop and invoke the image search tool by appending the string \texttt{<search><img></search>} at the end of your response. This will trigger a Google Lens search using the original image to retrieve relevant information that can help you confirm the visual content. \\
Once you have sufficient visual understanding, combine it with the user's question and assess whether you can confidently answer. If so, answer the question directly using your own knowledge. If not, invoke the text search tool by generating a concise and specific query, and output it in the format \texttt{<text\_search>your query here</text\_search>} at the end of your response. Carefully craft your query to accurately retrieve the information needed to help answer the question. The text search tool will then use Google Search to return relevant information based on your query. \\
You must include your reasoning inside \texttt{<reason>...</reason>} before taking any action, whether it is calling the image search tool, generating a text search query, or providing a final answer. The reasoning may involve analysis of the original image and question, interpretation of search results, or logical steps leading to the final answer. \\
All search results will be placed inside \texttt{<information>} and \texttt{</information>} and returned to you. When you are ready to answer the question, wrap your final answer between \texttt{<answer>} and \texttt{</answer>}, without detailed illustrations. For example: \texttt{<answer>Titanic</answer>}.
Here is the image and the question:
\texttt{<image>} 
\end{promptbox}

\begin{promptbox}{After Image Search Prompt}
\small
Please extract the visual element relevant to the user's question (such as faces, objects, locations, events, logos, or text) from the search results. Then, combine this information with the user's question and perform reasoning to determine whether a Google text search is needed to retrieve additional information to answer the question. \\
If a text search is needed, output the string \texttt{<text\_search>your query here</text\_search>} at the end of your response. Please generate a well-crafted query based on the visual element that will help retrieve the most relevant information. \\
If a text search is not needed, use your own knowledge to directly answer the user's question.
You must conduct your reasoning inside \texttt{<reason>} and \texttt{</reason>} before taking any action, whether it's generating a text search query or providing a final answer. If you decide to give the final answer, place it inside \texttt{<answer>} and \texttt{</answer>}, without detailed explanation or illustration.
For example: \texttt{<answer>Titanic</answer>}
\end{promptbox}

\begin{promptbox}{After Text Search Prompt}
\small
Please analyze the search results and the user's question and continue reasoning inside \texttt{<reason>} and \texttt{</reason>}. \\
If you determine that additional knowledge is still required to answer the user's question, stop responding to the question and instead report a warning by outputting the string 'Unable to answer due to lack of relevant information' at the end of your response. \\
If no further external information is needed, you should provide the final answer by placing it within \texttt{<answer>} and \texttt{</answer>}. The answer must be concise, clear, and to the point, without any additional explanation or elaboration.
\end{promptbox}

For our merged models, since their lineage traces back to Search-R1 \citep{searchr1}, we adopt Search-R1’s prompt convention by replacing \texttt{<reason>} with \texttt{<think>} and \texttt{<text\_search>} with \texttt{<search>}.

 The prompt used by Qwen-VL-DA and Qwen-VL-RAG are exactly the same as "Prompt used for Direct Answer" and "Prompts used for RAG workflow" in MMSearch-R1 \citep{mmsearchr1}, respectively.

\section{Search Tools}
\label{app:search_tool}
Following MMSearch-R1 \citep{mmsearchr1}, we implement two search tools: a text search tool and an image search tool.  
When the model invokes the image search tool, it retrieves the top-$k$ visually relevant web pages and returns a sequence of interleaved thumbnails and corresponding titles extracted from those pages.  
When the model invokes the text search tool with a textual query, it retrieves the top-$k$ relevant web pages and returns a summary of each page’s content along with its title.  
For further details, please refer to MMSearch-R1 \citep{mmsearchr1}.

\section{Hyper-parameters}
\subsection{Model Merging}
\label{app:merging_hyperparameters}
When applying the TA method \citep{ta} for model merging, we searched for the task vector coefficient~$\lambda$ within the set~$\{0.3, 0.5, 0.7\}$, ultimately selecting~$\lambda = 0.7$ for Qwen-VL and~$\lambda = 0.3$ for Search-R1.

For methods that inherently incorporate sparsification—TIES \citep{ties}, DARE \citep{dare}, and OBM—we fixed~$\lambda = 1$ and instead tuned the parameter density~$p$, which controls the fraction of top-$p$ most important parameters to retain. We evaluated~$p$ over the values~$\{0.3, 0.5, 0.7, 0.9\}$, and selected~$p = 0.7$ for both TIES and OBM, and~$p = 0.9$ for DARE.

\subsection{Training}
\label{app:training_details}
Our training hyper-parameters are identical to those used in MMSearch-R1~\citep{mmsearchr1}. Please refer to that work for implementation details.

\section{SimpleVQA}
\label{app:simplevqa}

As illustrated in Table \ref{tab:simplevqa}, the specially trained model actually achieved lower accuracy than Qwen-VL, which directly generates answers without search. Moreover, the RAG pipeline with a 100\% search rate further degraded performance, suggesting that SimpleVQA may not be an information-seeking dataset and is therefore unsuitable for evaluating the capabilities of multimodal search agents in this work.

\begin{table}[ht]
\centering
\caption{Performance on SimpleVQA.}
\label{tab:simplevqa}
\small % 或 \footnotesize，根据实际效果选择
\begin{tabular}{lcc}
\hline
\textbf{Model} & \textbf{Acc} & \textbf{SR} \\ \hline
Qwen-VL-DA     & 26.75        & 0.00        \\
Qwen-VL-RAG    & 19.15        & 100.00      \\
MMSearch-R1    & 24.58        & 45.80       \\ \hline
\end{tabular}
\end{table}

%%%%%%%%%%%%%%%%%%%%%%%%%%%%%%%%%%%%%%%%%%%%%%%%%%%%%%%%%%%%%%%%%%%%%%%%%%%%%%%
%%%%%%%%%%%%%%%%%%%%%%%%%%%%%%%%%%%%%%%%%%%%%%%%%%%%%%%%%%%%%%%%%%%%%%%%%%%%%%%

\end{document}